# Motif Detection Inspired by Immune Memory

William Wilson, Phil Birkin and Uwe Aickelin

School of Computer Science, University of Nottingham, UK
wow,pab,uxa@cs.nott.ac.uk

**Abstract.** The search for patterns or motifs in data represents an area of key interest to many researchers. In this paper we present the Motif Tracking Algorithm, a novel immune inspired pattern identification tool that is able to identify variable length unknown motifs which repeat within time series data. The algorithm searches from a neutral perspective that is independent of the data being analysed and the underlying motifs. In this paper we test the flexibility of the motif tracking algorithm by applying it to the search for patterns in two industrial data sets. The algorithm is able to identify a population of meaningful motifs in both cases, and the value of these motifs is discussed.

## 1 Introduction

The investigation and analysis of time series data is a popular and well studied area of research. Common goals of time series analysis include the desire to identify known patterns in a time series, to predict future trends given historical information and the ability to classify data into similar clusters. These processes generate summarised representations of large data sets that can be more easily interpreted by the user.

Historically, statistical techniques have been applied to this problem domain. However, the use of Immune System inspired (IS) techniques in this field has remained fairly limited. In our previous work [20] an IS approach was proposed to identify patterns embedded in price data using a population of trackers that evolve using proliferation and mutation. This early research proved successful on small data sets but suffered when scaled to larger data sets with more complex motifs. In this paper we describe the Motif Tracking Algorithm (MTA), a deterministic but non-exhaustive approach to identifying repeating patterns in time series data, that directly addresses this scalability issue.

The MTA represents a novel Artificial Immune System (AIS) using principles abstracted from the human immune system, in particular the immune memory theory of Eric Bell [1]. Implementing principles from immune memory to be used as part of a solution mechanism is of great interest to the immune system community and here we are able to take advantage of such a system. The

MTA implements the Bell immune memory theory by proliferating and mutating a population of solution candidates using a derivative of the clonal selection algorithm [6].

A subsequence of a time series that is seen to repeat within that time series is defined as a motif. The objective of the MTA is to find those motifs. The power of the MTA comes from the fact that it requires no prior knowledge of the time series to be examined or what motifs exist. It searches in a fast and efficient manner and the flexibility incorporated in its generic approach allows the MTA to be applied across a diverse range of problems.

Considerable research has already been performed on identifying *known* patterns in time series [13]. In contrast little research has been performed on looking for *unknown* motifs in time series. This provides an ideal opportunity for an AIS driven approach to tackle the problem of motif detection, as a distinguishing feature of the MTA is its ability to identify *variable length unknown* patterns that repeat in a time series. In many data sets there is no prior knowledge of what patterns exist, so traditional detection techniques are unsuitable. In this paper we test the generic properties of the MTA by applying it to two industrial data sets and asses its ability to find variable length unknown motifs in that data.

The paper is structured as follows, Section 2 provides a discussion of the work performed in the field of motif detection. An explanation of the inspiration for the MTA is presented in Section 3 before various important terms and definitions are introduced and the pseudo code for the MTA is described in Section 4. Section 5 presents the results of the MTA when applied to the two industrial data sets. Section 6 describes the relevance of the MTA and suggests potential applications before moving on to conclude in Section 7.

## 2 Related Work

The search for patterns in data is relevant to a diverse range of fields, including biology, business, finance, and statistics. Work by Guan [10] and Benson [3] addresses DNA pattern matching using lookup table techniques that exhaustively search the data set to find recurring patterns. Investigations using a piecewise linear segmentation scheme [11] and discrete Fourier transforms [8] provide examples of mechanisms to search a time series for a particular motif of interest. Work by Singh [17] searches for patterns in financial time series by taking a sequence of the most recent data items and looks for re-occurrences of this pattern in the historical data. An underlying assumption in all these approaches is that the pattern to be found is known in advance. The matching task is therefore much simpler as the algorithm just has to find re-occurrences of that particular known pattern. The search for unknown motifs is much harder problem as no prior knowledge of the motif is available. The MTA was created to address this specific challenge.

The search for unknown motifs is at the heart of the work conducted by Keogh et al. Keogh's probabilistic algorithm[4], used as a comparison to the MTA in Section 5.1, extracts subsequences from the time series using the Symbolic Aggregate Approximation (SAX) technique. It then hashes the subsequences into buckets. Buckets with multiple entries represent potential motif candidates. The sections of the time series corresponding to these subsequences are examined to identify genuine motifs. Keogh's Viztree algorithm[12] uses the SAX technique to generate a set of symbol strings corresponding to sequences from the time series. These symbol strings are filtered into a suffix tree, where each branch corresponds to alternative symbol combinations. The suffix tree provides a visual illustration of the motifs present as the frequency of a motif is shown by the width of the tree branch.

Keogh's Probabilistic and Viztree algorithms are very successful in identifying motifs but they require additional parameters compared to the MTA. They also assume prior knowledge of the length of the motif to be found. Motifs longer and potentially shorter than this predefined length may remain undetected in full. In addition, incorporating assumptions regarding the motif length would appear to contradict the definition of a truly unknown motif. Work by Tanaka [18] attempts to address this issue by using minimum description length to discover the optimal length for the motif. Fu et al. [9] use self-organising maps to identify unknown patterns in stock market data, by representing patterns as perceptually important points. This provides an effective solution but again the patterns found are limited to a predetermined length.

An alternative approach is seen in the TEIRESIAS algorithm [16] which identifies patterns in biological sequences. TEIRESIAS finds patterns of an arbitrary length by isolating individual building blocks that comprise the subsets of the pattern. These are then combined into larger patterns. The methodology of building up motifs by finding and combining their component parts is at the heart of the MTA. To achieve this the MTA takes an IS approach, evolving a population of trackers that is able to detect motifs whilst making fewer assumptions about the data set and the potential motifs. It focuses on the search for unknown motifs of an arbitrary length, leading to a novel and unique solution inspired by the developmental stages leading to immunological memory in the human immune system.

## 3 Long and short term memory

The flexible learning approach of the human immune system is attractive as an inspiration, but without an adequate memory mechanism knowledge gained from the learning process would be lost. Memory represents a key factor in the success of the immune system. A difficulty arises in implementing a computational immune memory mechanism however, because very little is known about the biological mechanisms underpinning memory development[23]. Theories such as antigen persistence and long lived memory cells[15], idiotypic networks[5], and the homeostatic turnover of memory cells[24] have all attempted to explain the development and maintenance of immune memory. However, all have been contested. In contrast the attraction of the immune memory theory proposed by

Eric Bell is that it provides a simple, clear and logical explanation of memory cell development. This theory highlights the evolution of two separate memory pools, 'memory primed' and 'memory revertant'[2].

The human immune system represents a successful recognition tool. It must be able to quickly identify novel bacteria and viruses present in the system so that it can react accordingly and retain knowledge of those encounters for future reference. The presence of such a bacterial threat causes naive immune cells to become activated. This activation causes a rapid increase in cell numbers, termed proliferation. The rapidly expanding population of activated cells forms the short lived memory primed pool. The purpose of this growing pool is to increase the repertoire of the population. New cells created during proliferation undergo mutation in order to diversify from their parents. The cell population evolves in order to match potential variations in the bacteria that stimulated their parents. A form of pattern matching is being anticipated by the system. The activated cells circulate throughout the system and eliminate any bacteria that they interact with.

The high death rate of memory primed cells means most will die during circulation, however a small minority do survive and return to reach a memory revertant state. These cells reduce their excessive activation levels, becoming more stable, thereby lengthening their lifespan. These unique cells are able to produce clones naturally to sustain knowledge of a bacterial experience over the long term. These two distinct memory pools and the transfer mechanism between them, represent a key difference to other memory theories. This methodology provides the inspiration for memory development in our algorithm.

Through this approach one can see the immune system represents an ideal mechanism to address motif matching problems. It evolves a population of solution candidates in an attempt to match part of a novel pattern, it then mutates these successful population members so that matching solutions can be improved. More information regarding the inspiration behind the MTA can be found in [20].

In our novel algorithm the equivalent of the short term memory primed pool is generated using a derivative of the popular clonal selection algorithm[6] to proliferate all successfully matched candidates. This memory pool evolves through a process of directed proliferation and mutation, regulated through a process of controlled cell death. This rapidly expanding population provides a search mechanism that is able to investigate all solution alternatives quickly and effectively. Successful candidates from the short term memory pool transfer to the longer lived memory revertant pool. This long term memory pool is then used to permanently store records of the solutions found.

Having briefly introduced the inspiration for the MTA, the algorithm itself and a number of key terms and definitions used within the algorithm are defined in the following section.

## 4 The Motif Tracking Algorithm

Before we can present the pseudo code for the MTA we need to define some of the terms used by the algorithm:

**Definition 1**. *Time series*. A time series $T = t_1,...,t_m$ is a time ordered set of *m* real or integer valued variables. In order to identify patterns in *T* we break *T* up into subsequences of length *n* using a sliding window mechanism.

**Definition 2**. *Motif*. A subsequence from *T* that is seen to repeat at least once throughout *T* is defined as a motif. We use Euclidean distance to examine the relationship between two subsequences $C_1$ and $C_2$, $ED(C_1, C_2)$ against a match threshold *r*. If $ED(C_1, C_2) \leq r$ the subsequences are deemed to match and thus are saved as a motif. The motifs prevalent in a time series are detected by the MTA through the evolution of a population of trackers.

**Definition 3**. *Tracker*. A tracker represents a signature for a motif sequence that is seen to repeat. It has within it a sequence of 1 to *w* symbols that are used to represent a dimensionally reduced equivalent of a subsequence. The subsequences generated from the time series are converted into a discrete symbol string. The trackers are then used as a tool to identify which of these symbol strings represent a recurring motif. The trackers also include a match count variable to indicate the level of stimulation received during the matching process.

Based on the above definitions, below is the MTA pseudo code with a description of its main operations. We direct the readers attention to [22] for a more in depth description of this algorithm if further information is required. The parameters required in the MTA include the length of a symbol *s*, the match threshold *r*, and the alphabet size *a*.

*MTA Pseudo Code*

```
Initiate MTA (s, r, a)
Convert Time series T to symbolic representation
Generate Symbol Matrix S
Initialise Tracker population to size a
While ( Tracker population > 0 )
{
    Generate motif candidate matrix M from S
    Match trackers to motif candidates
    Eliminate unmatched trackers
    Examine T to confirm genuine motif status
    Eliminate unsuccessful trackers
    Store motifs found
    Proliferate matched trackers
    Mutate matched trackers
}
Memory motif streamlining
```

**Convert Time Series T to Symbolic Representation.** The MTA takes as input a univariate time series consisting of real or integer values. Taking the first order difference of $T$ we look at movements between data points allowing a comparison of subsequences across different amplitudes. To further minimise amplitude scaling issues we normalise the time series. In our previous work [20] the algorithm investigated motifs through consideration of each data point individually, creating a solution that was not scalable to larger data sets. In the MTA this problem is resolved as we investigate motifs by combining individual data points into sequences and comparing and combining those sequences to form motifs.

Keogh's SAX technique [4] is used to discretise the time series. SAX is a powerful compression tool that uses a discrete, finite symbol set to generate a dimensionally reduced version of a time series that consists of symbol strings. This intuitive representation has been shown to rival more sophisticated reduction methods such as Fourier transforms and wavelets [4].

Using this approach a window of size $s$ slides across the time series $T$ one point at a time. Each sliding window represents a subsequence from $T$. The MTA calculates the average of the values from the sliding window and converts it into a symbol string. The user predefines the size $a$ of the alphabet used to represent the symbols for the time series $T$. As $T$ has been normalised the breakpoints for each alphabet character can be identified as those that generate $a$ equal sized areas under the Gaussian curve [4]. The average value calculated for the sliding window is then examined against the breakpoints and converted into the appropriate symbol. This process is repeated for all sliding windows across $T$ to generate $m-s+1$ subsequences, each consisting of symbol strings comprising one character. Additional information on this process can be found in [22].

**Generate Symbol Matrix S.** The string of symbols representing a subsequence is defined as a **word**. Each word generated from the sliding window is entered into the symbol matrix $S$. The MTA examines the time series $T$ using these words and not the original data points to speed up the search process. Symbol string comparisons can be performed efficiently to filter out bad motif candidates, ensuring the computationally expensive Euclidean distance calculation is only performed on those motif candidates that are potentially genuine.

Having generated the symbol matrix $S$, the novelty of the MTA comes from the way in which each generation a selection of words from $S$, corresponding to the length of the motif under consideration, are extracted in an intuitive manner as a reduced set and presented to the tracker population for matching.

**Initialise Tracker Population to Size a.** The trackers are the primary tool used to identify motif candidates in the time series. A tracker comprises a sequence of 1 to $w$ symbols. The symbol string contained within the tracker represents a sequence of symbols that are seen to repeat throughout $T$.

Tracker initialisation and evolution is tightly regulated to avoid proliferation of ineffective motif candidates. The initial tracker population is constructed of size $a$ to contain one of each of the viable alphabet symbols predefined by the user. Each tracker is unique, to avoid unnecessary duplication.

Trackers are created of a length of one symbol and matched to motif candidates via the words presented from the stage matrix S. Trackers that match a word are stimulated and become candidates for proliferation as they indicate words that are repeated in T. Given a motif and a tracker that matches part of that motif, proliferation enables the tracker to extend its length by one symbol each generation until its length matches that of the motif.

**Generate Motif Candidate Matrix M from S.** The symbol matrix S contains a time ordered list of all words, each containing just one symbol, that are present in the time series T. Neighbouring words in S contain significant overlap as they were extracted via sliding windows. Presenting all words in S to the tracker population would result in inappropriate motifs being identified between neighbouring words. To prevent this issue such 'trivial' match candidates are removed from the symbol matrix S in a similar fashion to that used in [4].

Trivial Match Elimination (TME) is achieved as a word is only transferred from S for presentation to the tracker population if it differs from the previous word extracted. This allows the MTA to focus on significant variations in the time series and prevents time being wasted on the search across uninteresting variations.

Excessively aggressive TME is prevented by limiting the maximum number of consecutive trivial match eliminations to s, the number of data points encompassed by a symbol. In this way a subsequence can eliminate as trivial all subsequences generated from sliding windows that start in locations contained within that subsequence (if they generate the same symbol string) but no others. The reduced set of words selected from S is transferred to the motif candidate matrix M and presented to the tracker population for matching.

TME speeds up the search process as it greatly reduces the number of motif candidates compared. However, TME can result in motifs being missed during the search. To account for this the MTA can also be run with No Trivial Match Elimination (NTME). This generates a far more accurate search at the cost of longer execution times.

**Match Trackers to Motif Candidates.** During an iteration each tracker is taken in turn and compared to the set of words in M. Matching is performed using a simple string comparison between the tracker and the word. A match occurs if the comparison function returns a value of 0, indicating a perfect match between the symbol strings. Each matching tracker is stimulated by incrementing its match counter by 1.

**Eliminate Unmatched Trackers.** Trackers that have a match count $>1$ indicate symbols that are seen to repeat throughout T and are viable motif candidates. Eliminating all trackers with a match count $<2$ ensures the MTA only searches for motifs from amongst these viable candidates. Knowledge of possible motif candidates from T is carried forward by the tracker population. After elimination the match count of the surviving trackers is reset to 0.

**Examine T to Confirm Genuine Motif Status.** The surviving tracker population indicates which words in *M* represent viable motif candidates. However motif candidates with identical words may not represent a true match when looking at the time series data underlying the subsequences comprising those words. In order to confirm whether two matching words *X* and *Y*, containing the same symbol strings, correspond to a genuine motif we need to apply a distance measure to the original time series data associated with those candidates. The MTA uses the Euclidean distance to measure the relationship between two motif candidates *ED(X,Y)*.

If *ED(X,Y)* ≤ r a motif has been found and the match count of that tracker is stimulated. A memory motif is created to store the symbol string associated with *X* and *Y*. The start locations of *X* and *Y* are also saved. For further information on the derivation and selection of this matching mechanism please refer to [22].

The MTA then continues its search, focusing only on those words in *M* that match the surviving tracker population, in an attempt to find all occurrences of the potential motifs. The trackers therefore act as a pruning mechanism, reducing the potential search space to ensure the MTA only focuses on viable candidates.

**Eliminate Unsuccessful Trackers.** The MTA now removes any unstimulated trackers from the tracker population. These trackers represent symbol strings that were seen to repeat but upon further investigation with the underlying data were not proven to be valid motifs in *T*.

**Store Motifs Found.** The motifs identified during the confirmation stage are stored in the memory pool for review. Comparisons are made to remove any duplication. The final memory pool represents the compressed representation of the time series, containing all the re-occurring patterns found.

**Proliferate Matched Trackers.** Proliferation and mutation are needed to extend the length of the tracker so it can capture more of the complete motif. At the end of the first generation the surviving trackers, each consisting of a word with a single symbol, represent all the symbols that are applicable to the motifs in *T*. Complete motifs in *T* only consist of combinations of these symbols. Therefore, these trackers are stored as the mutation template for use by the MTA.

Proliferation and mutation to lengthen trackers will only involve symbols from the mutation template and not the full symbol alphabet, as any other mutations would lead to unsuccessful motif candidates. During proliferation the MTA takes each surviving tracker in turn and generates a number of clones equal to the size of the mutation template. The clones adopt the same symbol string as their parent.

**Mutate Matched Trackers.** The clones generated from each parent are taken in turn and extended by adding a symbol taken consecutively from the mutation template. This creates a tracker population with maximal coverage of all potential motif solutions and no duplication. This process forms the equivalent of the short term memory primed pool identified by Bell [2, 4].

The tracker pool is fed back into the MTA ready for the next generation. A new motif candidate matrix $M$ consisting of words with two symbols must now be formulated to present to the evolved tracker population. In this way the MTA builds up the representation of a motif one symbol at a time each generation to eventually map to the full motif using feedback from the trackers.

Given the symbol length $s$ we can generate a word consisting of two consecutive symbols by taking the symbol from matrix $S$ at position $i$ and that from position $i+s$. Repeating this across $S$, and applying trivial match elimination, the MTA obtains a new motif candidate matrix $M$ in generation two, each entry of which contains a word of two symbols, each of length $s$.

The MTA continues to prepare and present new motif candidate matrix data to the evolving tracker population each generation. The motif candidates are built up one symbol at a time and matched to the lengthening trackers. This flexible approach enables the MTA to identify unknown motifs of a variable length. This process continues until all trackers are eliminated as non matching and the tracker population is empty. Any further extension to the tracker population will not improve their fit to any of the underlying motifs in $T$.

**Memory Motif Streamlining.** The MTA streamlines the memory pool, removing duplicates and those encapsulated within other motifs to produce a final list of motifs that forms the equivalent of the long term memory pool.

## 5 Results

Here we examine the MTA's performance on two publicly available industrial data sets. The MTA was written in C++ and run on a Windows XP machine with a Pentium M 1.7 Ghz processor with 1Gb of RAM.

### 5.1 Steamgen Data

In this section the MTA is compared to the probabilistic motif detection algorithm developed by Keogh. Keogh's algorithm is described in detail in [4]. It has been used as a basis for many motif detection algorithms [12, 18] and provides an ideal comparison for the MTA. Keogh was kind enough to provide the source code for a simplified version of his probabilistic motif detection algorithm and this was used as our basis of comparison.

One change was made to Keogh's source code to enable a direct comparison with the MTA. Motif detection in this version of Keogh's algorithm only goes as far as matching the symbol strings of the sequences being compared. No subsequent comparison of the sequences underlying those symbol strings is performed to confirm the existence of the motif. To incorporate this confirmation stage the same Euclidean distance function used in the MTA, as described in [22], is incorporated into Keogh's code. This ensures only genuine motifs are reported.

The two algorithms are compared using the Steamgen data set. This is a publicly available data set [1] used for motif detection. The data was generated using fuzzy models [7] applied to the model of a steam generator at the Abbott Power Plant in Champaign [14]. The Steamgen data set consists of every tenth observation taken from the steam flow output information, starting with the first observation. This specific selection criterion was used by Keogh and has been followed here for the purposes of comparison. The Steamgen data set contains 960 items with significant amplitude variation and is illustrated in Figure 1. From an initial scan by eye it is unclear whether any significant motifs exist, representing an ideal challenge for the two algorithms.

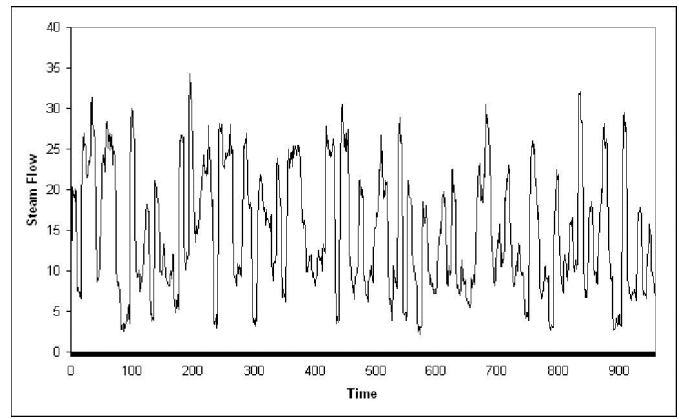

**Fig. 1.** A graphical representation of the Steamgen data set.

The parameters for the MTA for this data set have been selected based on the results of previous testing [22, 21]. An alphabet size *a=6* was set in accordance with that selected by Keogh. In addition the MTA uses a symbol size *s = 10*. A bind threshold of *r = 0.15* is selected to ensure only close fitting motifs are identified. For an analysis of the sensitivity of the MTA to changes in these parameters readers are directed to the analysis in our earlier work [22, 21].

Keogh's algorithm requires seven parameters to be established.

1. **Motif length**. Defines the length of the motif searched for.
2. **Number of symbols**. Defines the number of symbols used to represent the motif.
3. **Alphabet size**. Corresponds to the parameter *a* used in the MTA.
4. **Mask size**. Keogh's algorithm uses random projections to generate a collision matrix to identify potential motifs (see [4] for more information). The mask size determines how many symbols in the motif are used in the random projection.

---
[1] The data set can be accessed from http://homes.esat.kuleuven.be/~tokka/

**5. Projection iterations**. Defines how many times the random projection process is performed.

**6. Cut-off**. Defines the minimum threshold for the number of collisions in the collision matrix in order to identify if that sequence represents a motif candidate.

**7. Bind threshold**. Corresponds to the parameter *r* used in the MTA.

Keogh's algorithm requires that the length of the motif be specified in advance. The algorithm is then run for each of these different lengths. For the Steamgen data the motif length was set to 80, 70, 60, 50 and 40. To ensure the symbol length *s* is consistent across both algorithms the number of symbols per motif is calculated as the motif length divided by *s = 10*. An alphabet size of 6 is selected in line with the MTA. A mask size of 4 and a projection iteration of 20 were established by Keogh as suitable for this data set. Since we only want to identify good matching motifs the cut-off was set to 20. This implies only those motifs that match completely in all of the 20 random projections are identified as motifs. Finally a bind threshold of 0.15 is set in accordance with the MTA.

Keogh's algorithm is run with No Trivial Match Elimination NTME to provide a detailed search for motifs of the lengths specified. These represent the benchmark motifs that are compared to those found my the MTA.

Keogh's algorithm finds no motifs exist if the motif length is 80. A motif 'M1', consisting of two sequences at locations 65 and 873, is found when the motif length falls to 70. With a length of 60 two additional motifs are found. The first is located at positions 69 and 877 and the second is found at positions 77 and 885. These motifs represent partial subsets of M1 and do not reflect new motifs in the data. A motif length of 50 uncovers five different motifs, however, once again these all reflect partial subsets of M1. With a motif length of 40 Keogh's algorithm identifies 75 different motifs and some of these are distinct from M1.

However, many of the 75 motifs correspond to similar motif patterns that are offset. For example, one motif is found at location 79 and 887 while the next is found at 80 and 888. These sequences represent the same motif. In the MTA this issue is resolved by streamlining the memory pool. Such a mechanism is not available in Keogh's algorithm. Instead for the purposes of comparison the list of 75 motifs is manually condensed and the best fitting motif is kept to represent each motif pattern. All offset duplicates are eliminated. This reduces the number of unique motifs of length 40 from 75 to 15. These benchmark motifs are listed in Table 1.

In just one run the MTA generates a similar list of motifs. With NTME the MTA finds one motif of length 70, located as positions 67 and 875. This is consistent with motif M1 found by Keogh, offset by just two data points. The sequences representing this motif are illustrated in Figure 2. A dominant motif is clearly evident in the Steamgen data set and it has been found by Keogh's algorithm and the MTA.

The MTA finds eleven motifs with lengths varying between 40 and 50. These are directly comparable to the fifteen motifs found by Keogh's algorithm. A large number of other motifs of lengths less than 40 are also identified. These motifs are now compared to the fifteen benchmark motifs listed in Table 1. For example, benchmark motif 6 from Keogh's algorithm occurs at locations 569 and 785 with a length of 40. Motif 11 from the MTA occurs at location 563 and 779 with a length of 45. The MTA identifies the same approximate motif as Keogh except it is longer and offset by 6 data points. The process is repeated for the MTA with TME and for Keogh's algorithm with TME. The results are presented in Table 2.

**Table 1.** The motifs of length 40 detected in the Steamgen data set using Keogh's probabilistic algorithm. Locations of the motifs along with the degree of similarity are reported.

| Motif ID. | Location 1. | Location 2. | Euclidean Distance. |
|---|---|---|---|
| 1 | 79 | 887 | 0.92 |
| 2 | 52 | 356 | 1.40 |
| 3 | 392 | 641 | 1.94 |
| 4 | 232 | 296 | 2.22 |
| 5 | 227 | 339 | 1.98 |
| 6 | 569 | 785 | 2.06 |
| 7 | 10 | 234 | 1.63 |
| 8 | 1 | 337 | 2.29 |
| 9 | 98 | 450 | 2.03 |
| 10 | 366 | 754 | 2.12 |
| 11 | 525 | 917 | 1.93 |
| 12 | 386 | 802 | 2.31 |
| 13 | 464 | 896 | 2.07 |
| 14 | 454 | 910 | 2.22 |
| 15 | 153 | 769 | 2.43 |

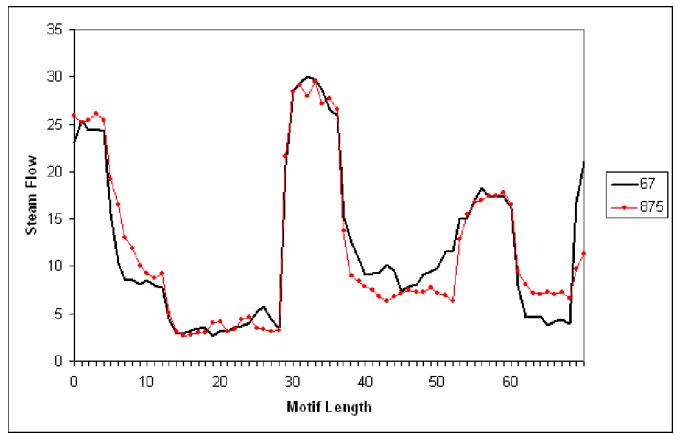

**Fig. 2.** The sequences corresponding to motif M1 in the Steamgen data, as identified by the MTA at locations 67 and 875

**Table 2.** A comparison of the motifs detected by Keogh's probabilistic algorithm against those found by the MTA, for the fifteen motifs identified in Table 1.

|       | MTA   |              | MTA   |              | Keogh |              |
|-------|-------|--------------|-------|--------------|-------|--------------|
|       | NTME  |              | TME   |              | TME   |              |
| Motif | Found | Length Error | Found | Length Error | Found | Length Error |
| 1     | Yes   | 0            | Yes   | 0            | Yes   | 0            |
| 2     | Yes   | 0            | Yes   | 0            | Yes   | 0            |
| 3     | Yes   | 0            | Yes   | -10          | Yes   | 0            |
| 4     | Yes   | 0            | Yes   | 0            | Yes   | 0            |
| 5     | Yes   | -10          | Yes   | -10          | Yes   | 0            |
| 6     | Yes   | 0            | Yes   | 0            | Yes   | 0            |
| 7     | Yes   | 0            | Yes   | 0            | No    | -40          |
| 8     | Yes   | -3           | Yes   | -3           | No    | -40          |
| 9     | Yes   | -10          | Yes   | -10          | No    | -40          |
| 10    | Yes   | -6           | No    | -40          | No    | -40          |
| 11    | Yes   | -10          | Yes   | -10          | No    | -40          |
| 12    | Yes   | 0            | Yes   | 0            | No    | -40          |
| 13    | Yes   | 0            | Yes   | 0            | No    | -40          |
| 14    | Yes   | 0            | Yes   | -10          | No    | -40          |
| 15    | Yes   | 0            | Yes   | -10          | No    | -40          |
|       |       | **-39**      |       | **-103**     |       | **-360**     |

Table 2 shows the MTA identifies all fifteen benchmark motifs when NTME is used. However, the length errors indicate that five of the motifs are shorter than the lengths established by Keogh's algorithm. For example, motif 8 is 37 data points long rather than 40. The flexibility of the MTA means that even though the full length of 40 is not recognised the MTA still recognises a significant proportion of these motifs. It can also identify motifs with lengths that are not divisible by the symbol size. Such features are not possible with Keogh's algorithm. In total the MTA missed 39 data points or 6.5% of the total length of all fifteen motifs.

With TME activated the MTA identifies fourteen of the fifteen benchmark motifs. Motif 10 is lost during the search. The length errors indicate that seven motifs are perfect matches to the benchmark motifs whilst the remainder represent only partial matches. The discrepancy in the length of the partially detected motifs amounts to 103 data points or 17.17% of the total motif length. In contrast Keogh's algorithm only finds six of the fifteen motifs if TME is used. Those six are found with 100% accuracy however all knowledge of the remaining motifs is lost. This occurs because of the severe nature of TME in Keogh's approach.

The MTA limits the maximum number of consecutive trivial matches to the length of the symbol *s*. This ensures neighbouring motifs with the same symbol representation are not eliminated and can be compared. In contrast Keogh's algorithm eliminates all consecutive trivial matches, causing motifs to be lost.

This experiment shows TME causes a deterioration in the detection accuracy of both the MTA and Keogh's algorithm. This is not surprising as TME trades off detection accuracy in order to improve the efficiency of the search. Efficiency is measured in terms of execution time and the number of times the algorithm has to return to the original time series to access subsequences for comparison during the motif confirmation stage. With NTME the MTA takes approximately 161 seconds to run and requires 157,785 data accesses. TME reduces execution time by 97.49% to 4.036 seconds and the number of data accesses by 88.22% to just 18,592. TME therefore creates a much faster, but less accurate solution.

Keogh's algorithm is considerably faster than the MTA if NTME is used. Each execution run only takes between two to three seconds. This is achieved due to the simpler search operation that is performed and use of Matlab rather than C++ to code the algorithm. This creates a big advantage for Keogh's solution as it can produce results nearly instantaneously. However, Keogh's algorithm assumes prior knowledge of the motif length and has to be rerun for different motif lengths. The MTA only needs to be run once.

To test the completeness of these results it is important to establish whether the MTA can find any motifs not found by Keogh's algorithm. Review of the MTA motifs highlights two that do not correspond to any of the benchmark motifs listed in Table 1. These motifs are listed in Table 3.

Table 3. Motifs in the Steamgen data set found by the MTA but missed by Keogh's detection algorithm.

| Motif | Symbolic Representation | Length. | Location 1 | 2 | Euclidean Distance |
|---|---|---|---|---|---|
| 1 | cbec | 40 | 218 | 418 | 1.95 |
| 2 | eccd | 40 | 275 | 546 | 2.57 |

Recalculating the Euclidean distance for each of these motif sequences confirms that they are valid motifs and should have been found by Keogh's approach. The only explanation for their omission is that the Euclidean distance calculation was never performed on these sequences. This would only occur if the sequences comprising the two motifs had different symbol strings and as a result were dismissed as non matching based on their symbolic representation.

To validate this theory the symbol strings relating to motifs 1 and 2 in Table 3 are re-assessed. Using the SAX technique [22] the sequences corresponding to motif 1 translate in Keogh's approach to the strings **ebce** for location 218 and **dbce** for location 418. The symbol for the first frame of each sequence differs. Even though the PAA values at the start of each sequence are similar

(0.433622 and 0.410164) they are located on either side of the boundary point 0.43 to generate a different symbol. Because of this Keogh's algorithm rejects the sequences as non matching and the motif is missed.

The same issue occurs for motif 2. Although the PAA values underlying the sequences are similar on two occasions they lie on either side of the boundary points. This generates different symbol strings (**cebd** and **bebe**) for the two sequences. The sequences are rejected by Keogh's algorithm as non matching and the motif is missed.

This issue does not occur for the MTA because it uses a different method to generate symbols. The MTA uses a global transformation. It normalises the whole time series first and then calculates the PAA values using a sliding window on that data. In contrast Keogh's algorithm uses a local transformation, normalising only across the sequences being compared. It appears in this case that normalising over a smaller range makes Keogh's approach more susceptible to inappropriate symbol representations, but this problem could also effect the MTA.

This highlights a weakness of the symbolic representation as closely matching sequences can be inappropriately dismissed if they lie close to the boundary points. A potential resolution is to allow neighbouring symbols to match each other as well as themselves. For example **b** could be allowed to match **a**, **b**, and **c**. This solution has already been adopted by Keogh [4].

However, this approach significantly increases the number of sequences that match in the symbol domain. Even with perfectly matching symbol sequences the underlying data can be quite disparate and reflect a non match. Increasing the match potential of symbol strings would dramatically increase the number of data accesses required in order to dismiss these sequences. This exponentially increases execution time. The benefit of a dimensionally reduced data set, derived from the symbolic representation, would be undermined. Furthermore, this matching mechanism would directly conflict with the proliferation and mutation mechanisms used in the MTA. A more appropriate solution would be to simply rerun the search with different alphabet sizes and the issue would be avoided.

### 5.2 Power Demand Data

In the second scenario the MTA is tested on the power demand data set [21]. This is a publicly available data set [1] containing 35,040 data values. Each data point represents the average power demand (in Kilo Watts) for each 15 minute interval during 1997 for the ECN research centre in the Netherlands [19].

Power demand rises when the research centre is open on week days but drops during evenings and weekends, as illustrated in Figure 3. The nature of these repeating cycles suggest that the data will contain a large number of motifs. Hypothetically motifs should exist for each working hour, each working day, each five day week, each working month and each weekend etc. These motifs should also occur with a consistent periodicity.

---
[1] The data can be obtained from www.cs.ucr.edu/~eamonn/TSDMA/

The MTA is used to find motifs in this power demand data and to identify inconsistencies in the periodicity of those motifs [21]. The inconsistencies could be indicative of potential anomalies in the data such as bank holidays or periods where the research centre had to close unexpectedly. These disturb the regularity of the motif sequence repeats and could be of value to a user of this data.

In 1997 the 28th and 31st of March were bank holidays in the Netherlands. As each data point equates to a fifteen minute interval these bank holidays relate to locations 8,352 and 8,640 in the data. To examine the ability of the MTA to identify these two bank holidays, 5,000 consecutive data items from location 5,000 to 10,000 were extracted as a subset for investigation by the MTA. This subset was selected rather than the whole data set because of the huge number of motifs that exist in the full data. Using the smaller subset we can test a proof of concept that the MTA can be used as a form of basic anomaly detection as well as a motif identification tool.

An alphabet size *a = 6* is selected based on results of previous testing [22, 21]. The symbol size is increased to *s = 250* due to the increased size of the power demand data set. A small match threshold of *r = 0.04* is selected to ensure the motifs are a close match for each other. An analysis of the sensitivity of the MTA to changes in these parameters is outside the scope of this particular analysis.

The performance of the MTA is assessed across ten execution runs. The motifs generated along with the average execution times and standard deviation for these runs are reported. The algorithm is run with Trivial Match Elimination (TME) and No Trivial Match Elimination (NTME) for comparison.

With NTME the MTA identifies seventeen motifs in the power demand data set. These motifs range in length from 250 to 1,373 data points. Ten of the seventeen motifs have a length equal to or greater than 500. Motifs shorter than 500 have been dismissed from the analysis for the sake of clarity, leaving the ten motifs seen in Table 4.

Table 4. Motifs detected in the Power Demand data set using MTA with NTME.

| Motif ID | Length | Frequency | Locations |
|---|---|---|---|
| 1 | 1,258 | 3 | 0. 1,339. 2,009. |
| 2 | 1,373 | 3 | 252. 1,598. 3,609. |
| 3 | 1,000 | 4 | 252. 1,598. 2,269. 3,608. |
| 4 | 1,005 | 3 | 926. 2,269. 3,608. |
| 5 | 973 | 4 | 0. 1,339. 2,010. 4,026. |
| 6 | 826 | 2 | 2,657. 3,423. |
| 7 | 598 | 7 | 0. 666. 1,338. 2,010. 2,681. 3,449. 4,025. |
| 8 | 500 | 7 | 327. 992. 1,665. 2,337. 2,912. 3,681. 4,366. |
| 9 | 519 | 6 | 452. 1,118. 1,791. 2,461. 3,805. 4,479. |
| 10 | 500 | 5 | 148. 818. 1,499. 2,162. 4,182. |

Table 4 shows a degree of overlap exists between the motifs. This is unsurprising given the nature of the power demand data. Motif 2 is the longest motif found by the MTA. It repeats three times across the data set and covers 4,092 (81.8%) of the 5,000 data points. The three occurrences of motif 2 (A, B and C) are illustrated in Figure 3.

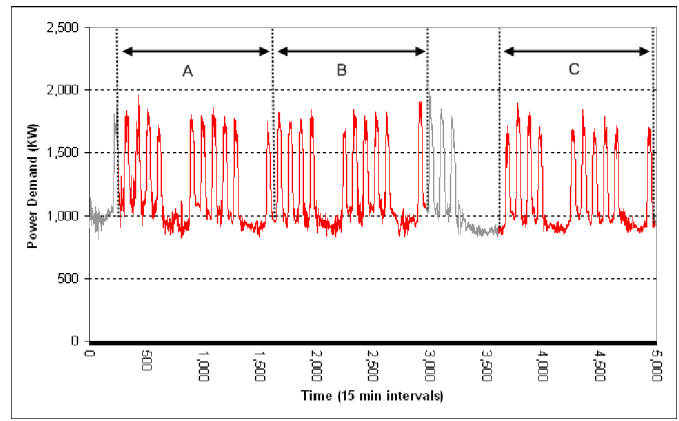

**Fig. 3.** Motif 2 from the power demand data set. The three repetitions of motif 2 are highlighted A, B and C.

Each occurrence of motif 2 lasts 1,373 data points which corresponds to approximately 2 weeks. The first two occurrences of motif 2 (A and B) occur back to back. Given this information one could hypothesise that the third occurrence (C) of motif 2 should occur directly after the completion of B. The hypothetical C should start near position 2,944 and end near position 4,317. However, the MTA finds C does not actually occur until point 3,609. The start of C is delayed by 665 data points (6.9 days). This indicates a potential anomaly exists in the data between points 2,944 and 4,317 as it was rejected as a match to motif 2. Apart from this region motif 2 encompasses the majority of the rest of the power demand data set.

Motif 7 has the highest frequency of all the motifs found by the MTA. It occurs seven times in the data set with a length of 598 data points (6.2 days). The seven occurrences of motif 7 labelled A to G are illustrated in Figure 4. Each occurrence of motif 7 encompasses two days where no power was required followed by four working days when the research centre required power. This motif is seen to occur regularly and covers 83.3% of the 5,000 data points.

Reviewing the intervals between each of the occurrences of motif 7 provides valuable insight into the data. B occurs 666 data points or 6.9 days after the start of A. The intervals between B, C, and D are also approximately seven days. This indicates that motif 7 occurs on a weekly basis. However, the interval between E and F is eight days, whilst that between F and G is only six days. The delay of

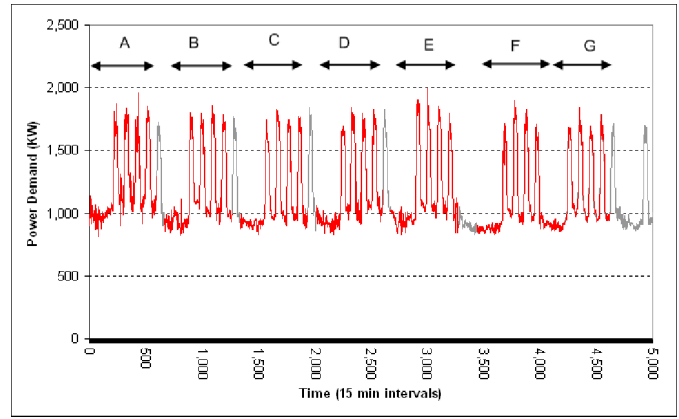

**Fig. 4.** Motif 7 from the power demand data set. The seven repetitions are highlighted A to G.

F by one day causes a disruption in the consistent periodicity of motif 7. If the periodicity of one week had been maintained F would have occurred from point 3,352 to 3,949. This suggests that an anomaly exists in the region 3,352 to 3,949 as it was rejected as a match to motif 7.

As each data point represents a fifteen minute interval the anomalous regions highlighted by motifs 2 and 7 translate to actual dates during 1997. Motif 2 suggests an anomaly occurred between the 23rd of March and the 7th of April. Motif 7 narrows this period to between the 28th of March and the 3rd of April. Earlier we stated two bank holidays did occur during these periods, on the 28th and on the 31st of March. The anomalies detected by the MTA are consistent with the bank holidays that occurred during 1997.

Reviewing Table 4 motif 6 is seen to relate to this anomalous period. Graphed in Figure 5 motif 6 encompasses the only two sequences (A and B) that correspond to the working weeks that include a bank holiday. A includes the bank holiday Friday on the 28th of March and B includes the bank holiday Monday on the 31st 1997. This shows that as well as identifying patterns that relate to normal working weeks, the MTA can also identify separate motifs for the weeks that include bank holidays.

The experiment is now re-run using Trivial Match Elimination TME and the MTA finds 16 motifs. To assess the impact of TME these motifs are compared to those in Table 4 using the following criteria and the results are presented in Table 5.

1. **Motif Found**. Is the motif found when TME is used?
2. **Frequency Error**. Records the discrepancy in the frequency of the motif found with TME compared to that found with NTME.

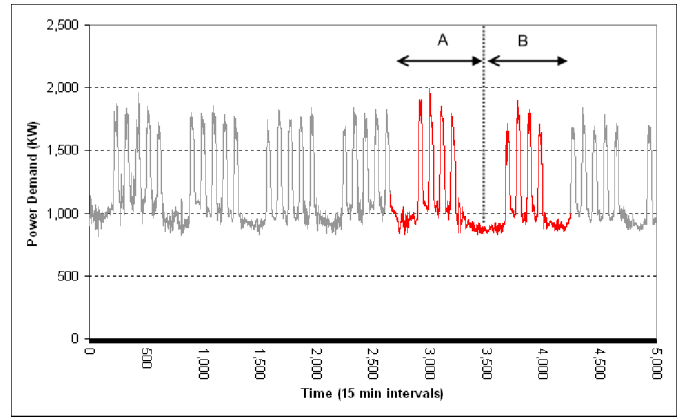

**Fig. 5.** Motif 6 from the power demand data set. The two repetitions of motif 6 are highlighted A and B.

  3. **Length Error**. Records the discrepancy in the length of the motif found with TME compared to that found with NTME. This is also reported as a percentage of the NTME motif length.
  4. **Location Error**. Records the maximum discrepancy in the location of the motif occurrences found with TME compared with NTME. This value is also reported as percentage of the NTME motif length.

**Table 5.** Impact on detection accuracy, with respect to the ten motifs from Table 4, by activating TME

| Motif | Found | Freq. Error | Length Error | | Location Error | |
|---|---|---|---|---|---|---|
| 1 | Yes | 0 | -8 | -0.6% | 10 | 0.8% |
| 2 | Yes | 0 | -11 | -0.8% | 23 | 1.7% |
| 3 | Yes | 0 | -250 | -25.0% | 62 | 6.2% |
| 4 | Yes | 0 | -255 | -25.4% | 62 | 6.2% |
| 5 | Yes | 0 | -223 | -22.9% | 10 | 1.0% |
| 6 | Yes | 0 | -37 | -4.5% | -4 | -0.5% |
| 7 | Yes | 0 | -98 | -16.4% | 11 | 1.8% |
| 8 | Yes | 0 | 0 | 0.0% | 30 | 6.0% |
| 9 | Yes | -2 | +231 | 44.5% | 25 | 4.8% |
| 10 | Yes | 0 | 0 | 0.0% | 61 | 12.2% |

With TME the MTA identifies all ten of the motifs found with NTME. However, the detection accuracy of the MTA suffers with TME. TME causes the MTA to miss two repetitions of motif 9. The correct frequency is identified for the remaining nine motifs.

The length errors in Table 5 indicate that, in seven of the ten cases, TME causes a shortening in the identified motif length. The most significant losses relates to motifs 3, 4 and 5, all of which were shortened by approximately 25% (220 to 250 data points). The remaining losses appear relatively minor and in one case TME results in a lengthening of the motif found. Overall however the motifs found using TME are only partial subsets of those found using NTME.

In nine out of ten cases the location errors show the motif starting positions found with TME are located after those found with NTME. The MTA misses the earlier start location, causing the motif length to fall. The sequences do reflect the same motif but the start locations are offset.

The errors in frequency, length and location indicate that TME causes a decline in the detection accuracy of the MTA. This is to be expected given TME sacrifices detection accuracy in order to speed up the search. Table 6 presents the average execution time and the number of accesses that are required back to the original time series for the ten runs with NTME and with TME. The standard deviation for those times is also highlighted to three decimal places.

Table 6. Efficiency comparison of NTME and TME using the power demand data set

|   | AvExecution Time | | Data Accesses |
|---|---|---|---|
|   | Seconds | S.D. |   |
| NTME | 3,639.87 | 241.327 | 4,885,096 |
| TME | 11.78 | 1.748 | 25,417 |

The MTA takes just 11.8 seconds to run with TME compared to over an hour with NTME. TME reduces the number of data accesses from over 4.8 million to just 25,417. Processing time is reduced by 99.68% while the number of data accesses is reduced by 99.48% with TME.

Given these efficiency improvements the lower detection accuracy for TME may be acceptable depending on the preferences of the user. The facility to use either TME or NTME provides the user with two alternative approaches that trade off accuracy versus efficiency, offering a flexible solution to the motif detection problem.

## 6 MTA Applications

The MTA represents a novel, abstract algorithm to identify unknowns motifs in a time series dataset in an intuitive and efficient manner. The population of motifs generated by the MTA is a potentially very useful resource that other algorithms could easily take advantage of.

Clustering and wavelet algorithms are usually seeded with random data upon initialisation. An alternative approach would be to seed these algorithms with the motif population generated by the MTA. The motifs represent known patterns that re-occur in the data set therefore giving the algorithms a head start in their analysis.

The motif generation process also represents a unique compression mechanism. The original time series is compressed to a reduced set representing the recurrent patterns in the data. This reduced set may be sufficient to provide a simple visual summary of the full data set.

Technical analysts working in the stock market use known and accepted patterns when analysing stock market performance. Such patterns include the 'head and shoulders' and 'cup and handle' patterns. The MTA could be applied to such stock market data and the motifs generated compared to these well established patterns. This analysis may be able to highlight other patterns that may also be of interest alongside those that are generally accepted.

A key potential application of the MTA would be to act as a support tool for forecasting. Executing the MTA on historical data would generate a population of motifs. When live data is received this information could represent a partial motif. Using principles from natural language one could compare the partial motif against the motif population and hypothesise the future direction or value for the data. For example, suppose the MTA was run on historical stock price data. The resulting motifs reflect reoccurring patterns in that data. Supposing the last five days of live data translated to the word **e**. Analysis of all the motifs that include the symbol **e** could then be used to assess the probability that a particular trend is reoccurring. This could provide additional information to assess which direction the market is heading.

# 7 Conclusion

Motifs and patterns are key tools for use in data analysis. By extracting motifs that exist in data we gain some understanding as to the nature and characteristics of that data. The motifs provide an obvious mechanism to cluster, classify and summarise the data, placing great value on these patterns. Whilst most research has focused on the search for *known* motifs, little research has been performed looking for *variable length unknown* motifs in time series. The MTA takes up this challenge, building on our earlier work to generate a novel immune inspired approach to evolve a population of trackers that seek out and match motifs present in a time series. The MTA uses a minimal number of parameters with minimal assumptions and requires no knowledge of the data examined or the underlying motifs, unlike other alternative approaches. Previous issues of scalability were addressed by using a discrete, finite symbol set to generate a dimensionally reduced version of the time series for investigation.

The MTA was evaluated using two industrial data sets and the algorithm was able to identify a motif population for each. Tests involving the steamgen data showed the MTA could identify a motif population that approximates closely to the motifs found using a derivative of the motif detection algorithm developed by Keogh. It was also able to find motifs not found by this alternative approach.

From the power demand data set the MTA was able to identify intuitive motifs that provided valuable insight into the nature of the data set. Irregularities in the motif periodicity indicated the existence of anomalies in the data. These correlated with the existence of bank holidays which disrupted the regularity of the motifs. In addition the MTA was able to successfully identify separate motifs corresponding to the bank holidays and those relating to normal working weeks.

The ability to identify anomalies in motifs that occur with a consistent periodicity could have direct relevance for numerous other applications. A prime example includes medical applications which record and monitor heart rates and breathing patterns. The ability to find anomalies in this type of data could prove extremely beneficial.

In both data sets the use of trivial match elimination was seen to reduce the detection accuracy of the MTA as discrepancies in the motif length, location and frequency are introduced. This deterioration is compensated for by a significant improvement in overall efficiency as the execution time and the number of data accesses are dramatically reduced if TME is used.

The ability to identify a population of truly "unknown" and meaningful motifs using minimal assumptions and a limited number of parameters ensures the the MTA offers a valuable contribution to an area of research that at present has received surprisingly little attention.